\documentclass[letterpaper]{article} 
\usepackage{aaai25}  
\usepackage{times}  
\usepackage{helvet}  
\usepackage{courier}  
\usepackage[hyphens]{url}  
\usepackage{graphicx} 
\usepackage{amsmath} 
\usepackage{multirow}
\usepackage{amssymb}
\usepackage{booktabs} 
\usepackage{xcolor}
\usepackage{natbib}  
\usepackage{caption} 
\usepackage{algorithm}
\usepackage{algorithmic}

\usepackage{newfloat}
\usepackage{listings}
\DeclareCaptionStyle{ruled}{labelfont=normalfont,labelsep=colon,strut=off} 
\floatstyle{ruled}
\newfloat{listing}{tb}{lst}{}
\floatname{listing}{Listing}
\title{Multi-scale Activation, Refinement, and Aggregation: Exploring Diverse Cues for Fine-Grained Bird Recognition}
\author{
	Zhicheng Zhang\textsuperscript{\rm 1},
	Hao Tang\textsuperscript{\rm 1,2},
	Jinhui Tang\textsuperscript{\rm 1,}\thanks{Corresponding author.}
}

\affiliations{
    \textsuperscript{\rm 1}~Nanjing University of Science and Technology, China\\
    \textsuperscript{\rm 2}~Centre for Smart Health, The Hong Kong Polytechnic University\\
    \{zzc\_666, jinhuitang\}@njust.edu.cn, howard.haotang@gmail.com
}

\usepackage{bibentry}

\begin{document}

\maketitle

\begin{abstract}
Given the critical role of birds in ecosystems, Fine-Grained Bird Recognition (FGBR) has gained increasing attention, particularly in distinguishing birds within similar subcategories. Although Vision Transformer (ViT)-based methods often outperform Convolutional Neural Network (CNN)-based methods in FGBR, recent studies reveal that the limited receptive field of plain ViT model hinders representational richness and makes them vulnerable to scale variance. Thus, enhancing the multi-scale capabilities of existing ViT-based models to overcome this bottleneck in FGBR is a worthwhile pursuit. In this paper, we propose a novel framework for FGBR, namely Multi-scale Diverse Cues Modeling (MDCM), which explores diverse cues at different scales across various stages of a multi-scale Vision Transformer (MS-ViT) in an ``Activation-Selection-Aggregation'' paradigm. Specifically, we first propose a multi-scale cue activation module to ensure the discriminative cues learned at different stage are mutually different. Subsequently, a multi-scale token selection mechanism is proposed to remove redundant noise and highlight discriminative, scale-specific cues at each stage. Finally, the selected tokens from each stage are independently utilized for bird recognition, and the recognition results from multiple stages are adaptively fused through a multi-scale dynamic aggregation mechanism for final model decisions. Both qualitative and quantitative results demonstrate the effectiveness of our proposed MDCM, which outperforms CNN- and ViT-based models on several widely-used FGBR benchmarks.
\end{abstract}

\section{Introduction}
Birds play a crucial role in the biological chain.
However, the advancement of industrialization and consequent environmental degradation have led to a sharp decline in numerous bird species.
Statistics \cite{malhotra2022habitat} indicate that nearly half of all bird species are in decline. Due to their sensitivity to  environmental changes, birds' activities can serve as indicators of environmental changes. Fine-Grained Bird Recognition (FGBR) has become a research hotspot, crucial for the conservation of bird species and natural habitats.

\begin{figure}[t!]
  \centering
  \includegraphics[width=\linewidth]{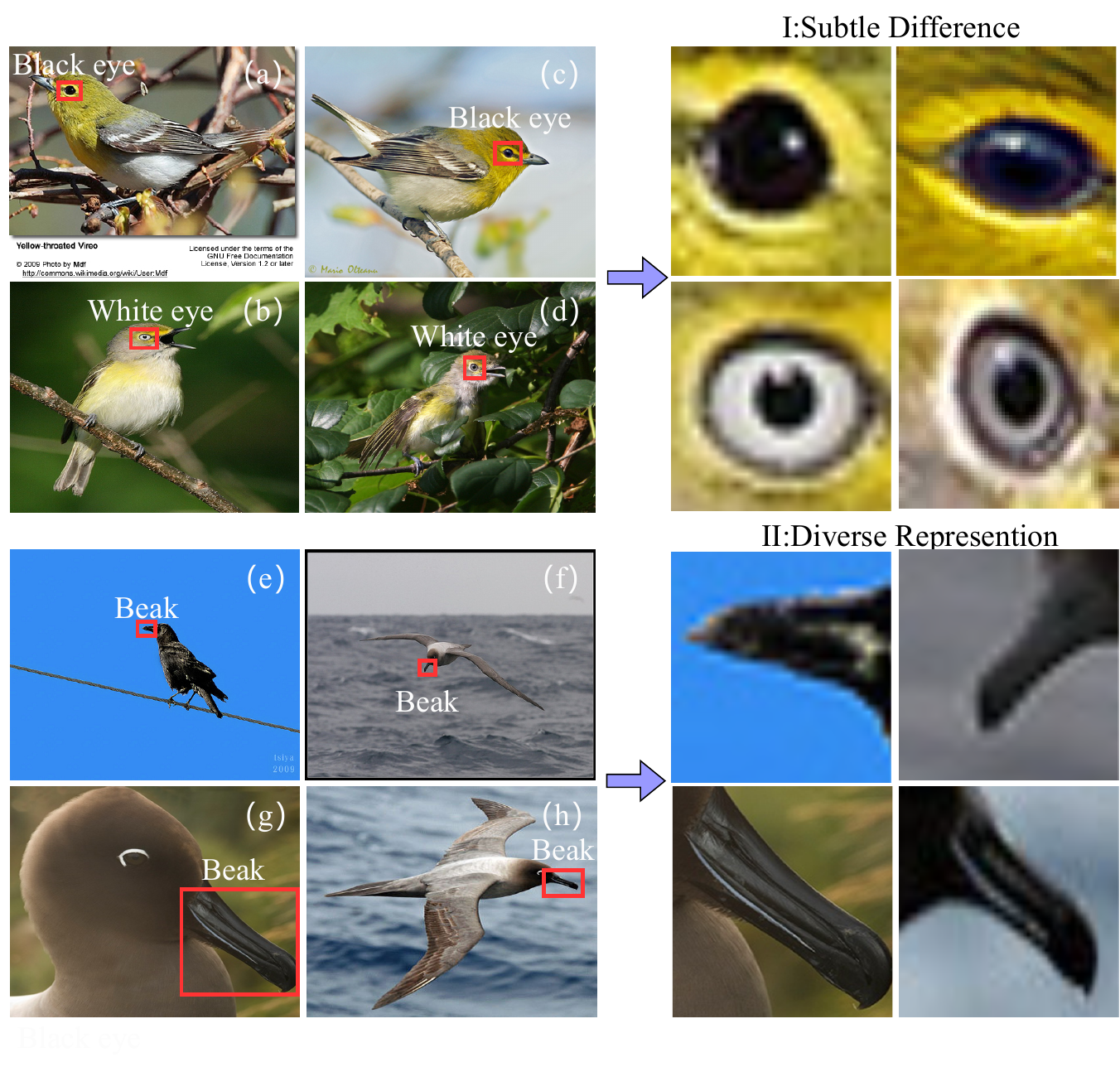}
  \caption{
    The primary challenges in FGBR are evident from bird images. Figures (a) to (d) show subtle species differences, often obscured by complex backgrounds. Figures (e) to (h) highlight significant scale variations between distant and close-up shots, making the same body parts appear different and complicating the recognition task.
    }\label{challenge}
\end{figure}

Fine-grained bird recognition is widely recognized as a challenging task due to the intrinsic characteristics of bird images~\cite{TangYLT22,JiangTL24}. These challenges can be summarized as follows:
\begin{itemize}
    \item \textbf{Complex Backgrounds:}
Bird habitats, whether in the wild or urban environments, often characterized by complex and variable backgrounds that can obscure the subtle features distinguishing one species from another~\cite{ZhaTST23}. For example, as shown in Figure~\ref{challenge}, the key difference between the Yellow-Throated Vireo and the White-Eyed Vireo lies in the eye color. Complex backgrounds can misdirect the model's attention, causing it to focus on irrelevant background elements instead of critical features, leading to potential misclassification.

    \item \textbf{Scale Variations:}
Bird images often exhibit significant scale variations due to differences in shooting distances, resulting in varying sizes of the same body part (\emph{e.g.,~}beak and eyes) across images (Figure~\ref{challenge}). These multi-scale differences complicate accurate recognition, requiring the model to consistently identify features such as the shape and texture of the beak or the color of the eyes, regardless of the scale at which they appear. 
\end{itemize}

Recently, Vision Transformers (ViTs) \cite{DosovitskiyB0WZ21} have demonstrated impressive performance across various visual tasks~\cite{Jiang000L24,shen2023git}. ViT divides images into patches, projects these patches into tokens, and then feeds them into transformer blocks for image modeling. However, pure ViT architectures face challenges when applied to fine-grained recognition tasks~\cite{FangJTL24,10038499,fu2024cross}.
Firstly, ViT models often struggle to capture the discriminative body parts crucial for birds. To address this issue, researchers have developed methods that encourage the model to focus on object regions rather than background noise. These approaches include selecting specific image patches that contain relevant features \cite{HeCLKYBW22,WangY021} or designing mechanisms that emphasize discriminative cues \cite{10023961,ZhuKLLTS22}. Secondly, ViT's reliance on fixed-size image patches makes it less effective at capturing multi-scale features. To overcome this limitation, previous works have introduced multi-scale modules, which allow the model to observe images at different scales across multiple stages and use the features extracted from the final stage for downstream tasks \cite{LiuL00W0LG21,LiW0MXMF22,0007LYYL023,DongZYTT24}. However, significant scale variations in FGBR remain difficult to manage, as certain features learned at appropriate scales in earlier stages may be discarded during later computations, leading to the loss of valuable information.

Unlike previous works on FGBR that primarily exploit general bird features at a single scale, this work proposes a novel ``Activation-Selection-Aggregation'' paradigm. This approach captures diverse cues by extracting multi-scale features across various stages of a multi-scale Vision Transformer (MS-ViT). 
To obtain these diverse cues, we first extract features from multiple stages and introduce a multi-scale cue activation module during the forward process to adjust and enhance cue distinctiveness. Consistent with this architecture, we then design the multi-scale token selection mechanism, which leverages deep semantic information to guide token selection in shallow layers, thereby enhancing robustness.
With cues learned across multiple stages, we could achieve recognition results at various scales. However, due to significant scale variations in FGBR, recognition results at inappropriate scales can negatively impact overall accuracy. To address this, we implement a gating mechanism that dynamically aggregates the recognition results.
We validate the effectiveness of our method on two popular bird datasets and a large-scale species recognition dataset. Additionally, we also provide intuitive visualizations demonstrating how our approach mitigates background noise and offer quantitative analyses of the multi-scale cues aggregation's performance.

\section{Related Work}
The primary challenges of FGBR include high visual similarity among subclasses, complex and variable background environments, and multi-scale representations. Naturally, researchers have explored part localization to extract part-level features, performing feature interaction and alignment for classification. Early  approaches typically designed localization sub-networks for FGBR \cite{ZhangDGD14,LinSLJ15,ZhangXEHZEM16}, but these methods often required highly accurate part-level annotations for training, which limits its development. In response, later researchers shifted towards weakly supervised methods, such as Region Proposal Networks (RPN) and attention mechanisms \cite{TangLPT20,HeCLKYBW22,KeCCLG23,yan2023progressive,DongYTTZ24,tang2024divide}.  

However, the localization process  significantly increases computational cost. Given the crucial role of feature learning in visual tasks, researchers have sought to enhance feature learning to capture subtle differences. \cite{ShenSWJY22} introduces additional convolutional layers to assess the importance of different regions within the feature map, adjusting activation relationships to uncover rich cues. Similarly, \cite{SongY21} employs extra convolutional layers to determine the importance of the feature map, adjusts its activations to encourage the model to focus on more potential regions. Some other researchers focus on identifying the most salient features and leveraging them for interaction and recognition. \cite{HeCLKYBW22} selects the most important tokens before the last layer to encourage the model to focus on discriminative parts. \cite{WangY021} aggregates the important tokens from each transformer layer to capture discriminative features.  Similar approaches have been explored by \cite{HuJZHZH021,abs-2303-06442}. Additionally,  some methods have investigated pairwise feature interactions to derive a unified yet discriminative image representation from paired images of different species \cite{ZhuangW020,ZhuKLLTS22}.

\begin{figure*}
\centering
\includegraphics[width=\textwidth]{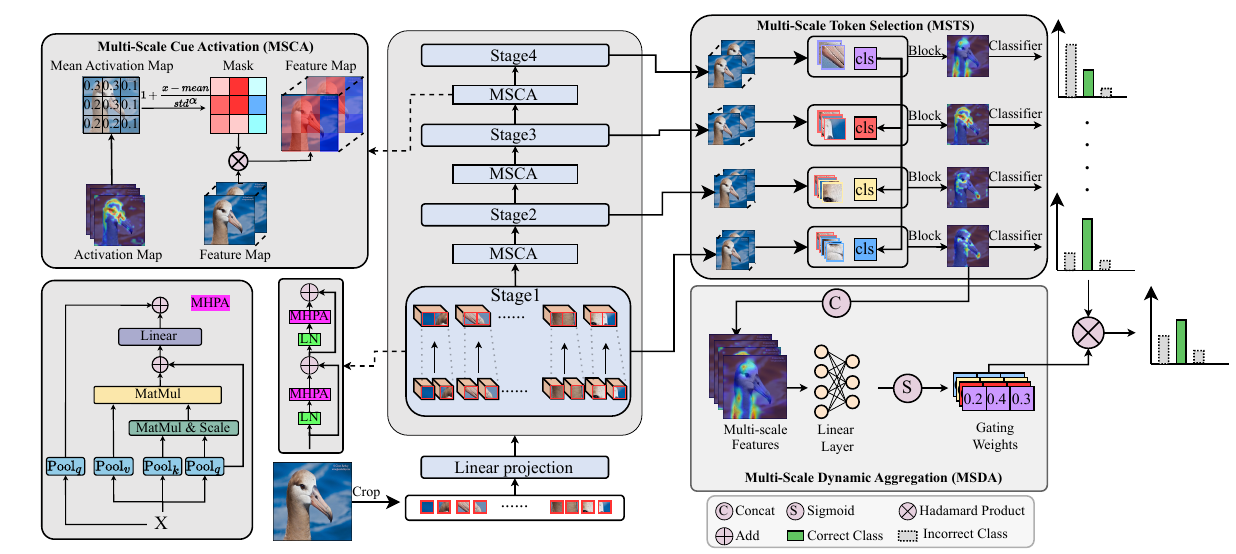}
\caption{The framework of our MDCM. During the forward pass, MSCA adjusts the activation of feature map to ensure the cues learned at different stage are mutually different. Subsequently, MSTS extracts diverse cues from multiple stages and filters out noisy regions. Finally, MSDA dynamic aggregates the  classification results for the final model decisions.}
\label{model}
\end{figure*}

\section{Multi-scale Diverse Cues Modeling}
Figure~\ref{model} illustrates the overall framework of our MDCM, which follows an ``Activation-Selection-Aggregation'' paradigm. 
 Firstly, the feature extractor is employed to capture cues at different scales across various stages. During the forward pass, we adjust the activation strength of these features to ensure that each stage learns distinct features. Secondly, to better focus on the salient parts of birds, we remove irrelevant background patches, highlighting regions crucial for classification. Finally, the multi-scale feature representations are fed into multiple classifiers to generate classification results, which are then aggregated through a gating mechanism for precise target classification.

\subsection{Multi-Scale Vision Transformer}
We utilize Multi-scale Vision Transformer (MS-ViT) \cite{LiW0MXMF22} as our primary backbone, due to its ability to capture multi-scale features from images effectively. Unlike traditional Vision Transformer~\cite{DosovitskiyB0WZ21}, which divides image into patches and project these patches into tokens, MS-ViT introduces a pooling operation within the multi-head self-attention mechanism, termed Multi-Head Pooling Attention (MHPA). This approach reduces the number of tokens, increases the number of channels, lowers the image resolution, and adjusts the scale. 

Let $X_0 \in \mathbb{R}^{h_{0} \times w_{0} \times c_{0}}$ represents the bird image, where $h_{0}$, $w_{0}$, and $c_{0}$ denote the height, width, and channels of the input image, respectively. Initially, the input image is divided into patches of size $o \times o$, which are then reshaped into one-dimensional patches sequence and passed through the patch embedding layer. A trainable class (\texttt{cls}) token is appended to the patch sequence, and a learnable position embedding is added to retain positional information. The embedding process is expressed as follows:
\begin{equation}
  \mathbf{Z} = [z^0, z^1 \mathbf{E}, z^2 \mathbf{E}, \cdots, z^N \mathbf{E}] + \mathbf{E}_{\text{pos}},
\end{equation}
where $\mathbf{E} \in \mathbb{R}^{(\frac{h_0 \cdot w_0}{o^2} \cdot c_0) \times D}$ is the patch embedding projection, $\mathbf{E}_{\text{pos}} \in \mathbb{R}^{(N+1) \cdot D}$ represents the position embedding, $N$ denotes the number of image patches, $z^0$  is the learnable \texttt{cls} token, and $z^n$ for $n \in \{1, 2, \ldots, N\}$ are the image patches. The MS-ViT consists of $L$ layers of Multi Head Pooling Attention (MHPA) and Multi-Layer Perceptron (MLP) blocks. The forward pass for the $l$-th layer is as follows:
\begin{align}
  \mathbf{Z}'_l &= \text{MHPA}(\text{LN}(\mathbf{Z}_{l-1})) + \mathbf{Z}_{l-1}, & l \in 1, 2, \ldots, L, \\
  \mathbf{Z}_l &= \text{MLP}(\text{LN}(\mathbf{Z}'_l)) + \mathbf{Z}'_l, & l \in 1, 2, \ldots, L,
\end{align}
where $\text{LN}(\cdot)$ denotes the layer normalization operation. MS-ViT utilizes the \texttt{cls} token of the last encoder layer, $z_L^0$, as the representation of the global feature, which is then forwarded to a classifier head for the final recognition.

\subsection{Multi-Scale Cue Activation}
If cues learned at different stages of MS-ViT are highly similar, it would impede the effective utilization of multiscale information in bird images. Inspired by the success of the attention-based erasing mechanism in CNN-based architectures~\cite{SongY21,ShenSWJY22}, we propose the parameter-free Multi-Scale Cue Activation (MSCA) module for MS-ViT. This module adjusts the activation of discriminative cues at each stage to explicitly learn multiple scale-specific representations, distinguishing it significantly from previous ViT-based methods.

During the forward propagation of the MS-ViT, the image is encoded into tokens for computation. Since the activation strength of cues should not be adjusted arbitrarily, it is crucial to first assess the importance of each token. Specifically, in MHPA, the tokens $z \in \mathbb{R}^{(N+1) \times D}$ are mapped and pooled into three matrices: $\mathbf{Q}$ (query), $\mathbf{K}$ (key), and $\mathbf{V}$ (value). The attention operation is performed as $\mathrm{Softmax}(\frac{\mathbf{Q}\mathbf{K}^\top}{\sqrt{d}})\mathbf{V}$, where $d$ is the dimension  of the query vector. It is widely known that the \texttt{cls} token in ViTs tends to focus more on class-specific tokens rather than on tokens associated with non-object regions. Therefore, we propose utilizing the attentiveness of the \texttt{cls} token to other tokens as an activation map to identify the most important cues. Specifically, the activation map $\mathbf{A}$ is defined as $\mathbf{A}=\mathrm{Softmax}(\frac{\mathbf{q}_{class}\cdot \mathbf{K}^\top}{\sqrt{d}})$, where $\mathbf{q}_{class}$ represents the query vector of the \texttt{cls} token, and $\mathbf{K}$ and $\mathbf{V}$ correspond to the key and value matrices, respectively. Consequently, the activation value $a_i$ (\emph{i.e.}, the $i$-th entry in $\mathbf{A}$) reflects the importance of the $i$-th token.

Some studies~\cite{SerranoS19,AbnarZ20} indicate that high-level attention information may not always align with the true importance of the input tokens. To enhance robustness, we average the activation maps from previous stages to form the attention map for the current stage. To address the shape mismatch caused by pooling operations, we employ interpolation to adjust the shapes accordingly. Subsequently, we map the importance scores to an enhancement and suppression mask to adjust the activation of cues without introducing additional parameters. Specifically, for each element $a^{k}$ in $\mathbf{A}$ with the with index $k \in \{1,2,\cdots,N\}$, we apply the following operation to derive the scaling mask:
\begin{equation}\label{z-score}
  m^k = \frac{a^k-\mathrm{mean}(\mathbf{A}')}{\mathrm{std}(\mathbf{A}')^\gamma},
\end{equation}
where $\mathbf{A}' \in \mathbb{R}^{N}$ represents all elements in $\mathbf{A}$ except the first one $a^0$, $\gamma$ is a hyperparameter that regularizes the degree of activation adjustment, and $\mathrm{mean}(\mathbf{A}')$ and $\mathrm{std}(\mathbf{A}')$ denote the mean and standard deviation of $\mathbf{A}'$. The resulting value, $m^k$, serves as the weight for the suppression and enhancement mask corresponding to the $k$-th token. We then apply the mask through element-wise multiplication with the tokens, yielding the activated tokens as $\mathbf{Z}' = \mathbf{Z} \odot \mathbf{M}$, where $\mathbf{M} = \{m^{1}, m^{2}, \cdots, m^N\}$.

\subsection{Multi-Scale Token Selection}
Considering that noisy tokens corresponding to complex background regions may misdirect the model’s attention, the Multi-scale Token Selection (MSTS) mechanism is proposed to filter out noisy tokens. 
Let $\mathbf{Z'}_i = \{z_i^0,z_i^1,\cdots,z_i^{h_i\cdot w_i} \} $ denote the output from stage $i$. First, the \texttt{cls} token $z_i^0$ is detached, and the remaining patch tokens $\mathbf{P} = \{z_i^1,z_i^2,\cdots,z_i^{h_i \cdot w_i}\}$ are reshaped from $\mathbb{R}^{h_i  \cdot w_i \times c_i}$ to  $\mathbb{R}^{h_i \times w_i \times c_i}$. Next, we perform patch merging on $\mathbf{P}$ and flatten it to a 1D token sequence $\mathbf{P}' \in \mathbb{R}^{N_i \times c_i}$, where $N_i = (\frac{h_i\cdot w_i}{4})$.  The patch merging method, inspired by Swin Transformer~\cite{LiuL00W0LG21}, expands the model's receptive field.

MSTS calculates scores $\mathbf{S}_i$ for tokens in $\mathbf{P}'$ to determine their importance and retains the top $k_i$ tokens. 
The importance of each token $z_i \in \mathbb{R}^{c_i}$ is determined by averaging its activation across channels. The score $s$ for each token is computed as:
\begin{equation}
  s = \frac{1}{c_i}\sum_{j=1}^{c_i}z(j),
\end{equation}
where $c_i$ is the number of channels. The scores $\mathbf{S}_i = \{s^1,s^2,\cdots ,s^{N_i}\}$ are sorted in descending order, and the top $k_i$ indices $\mathbf{I}_i$ are selected. The corresponding tokens are then gathered from $\mathbf{P}'$ as follows:
\begin{align}
  \mathbf{I}_i &= \operatorname{topkIndex}(\mathbf{S}_i;k_i), \\
  \hat{\mathbf{P}_i} &= \operatorname{gather}(\mathbf{P}',\mathbf{I}_i).      
\end{align}

To leverage the rich semantic information from deeper layers for localization, MSTS selects tokens in shallow layers correspond to those chosen in deep layers, along with layer-specific tokens. For MS-ViT, Stages 1 and 2 are designated as shallow layers, while Stage 3 serves as the deep layer. Stage 3 captures rich semantic information with its higher resolution map, enabling flexible token selection.
For the \texttt{cls} token, instead of using the original \texttt{cls} token from each stage, we employ the \texttt{cls} Token Transfer (CTT) method~\cite{abs-2308-02161} to leverage the rich semantic information from deeper layers. Specifically, a linear layer projects the \texttt{cls} token $z_4^0$ from the last stage to the \texttt{cls} tokens of preceding stages, as defined:
\begin{equation}
  {z}_i^{0} = \mathbf{W}_{i}^{1} \left(\mathrm{ReLU}\left(\mathrm{BN}\left(\mathbf{W}_{i}^{0} {z}_4^{0}\right)\right)\right).
  \label{eq:layer_output}
\end{equation}

Finally, the $z_i^0$ and $\hat{\mathbf{P}}_i$ are concatenated and passed through a Squeeze and Excitation network~\cite{HuSS18} to explore cross-channel interactions. They are then processed through standard Transformer Blocks independently to obtain multi-scale \texttt{cls} tokens $\hat{z}_i^{0}$ for classification:
\begin{align}
  \hat{\mathbf{Z}}_i &=  \operatorname{Block}\left(\operatorname{SEM}\left(\operatorname{Concat}\left(z_i^0, \hat{\mathbf{P}}_i\right)\right)\right),\\
  \eta_{i} &= \mathbf{W}_i(\hat{z}_i^{0}),
\end{align}
where $\hat{z}_i^{0}$ is the \texttt{cls} token in $\hat{\mathbf{Z}}_i$. 

\subsection{Multi-Scale Dynamic Aggregation}
One of the primary challenges in FGBR arises from the scale differences caused by close-up and long-distance shots, making it difficult to rely on a single scale for effective recognition. To address this, our method derives recognition results from multi-scale cues and aggregates them. However, simple summation aggregation indiscriminately combines results from all scales, potentially incorporating inappropriate or even harmful cues. To resolve this, we introduce a gating mechanism that selectively aggregates the most relevant results from multi-scale cues. Details of the mechanism are as follows:

First, we concatenate all \texttt{cls} token $\hat{z}_i^0$ to construct the global multi-scale features $\mathcal{MF}$ of the image. Simultaneously, we stack the recognition results from the four stages to obtain the global classification result $\mathcal{MC} \in \mathbb{R}^{4\times n}$, where $n$ is the number of the classes:
\begin{align}
  \mathcal{MF} &= \operatorname{Concat}(\hat{z}^0_{1},\hat{z}^0_{2},\hat{z}^0_{3},\hat{z}^0_{4}),  \\
  \mathcal{MC} &= \operatorname{Stack}(\eta_1,\eta_2,\eta_3,\eta_4).          
\end{align}

Next, we compute a set of gating weights $\mathbf{G} \in \mathbb{R}^{4\cdot n}$ to guide the aggregation process:
\begin{equation}\label{gatingweight}
  \mathbf{G} = \sigma(\mathbf{W}\mathcal{MF}+b),
\end{equation}
where $\mathbf{W}$ and $b$ are learnable transformation weights and biases, and $\sigma$ is the element-wise sigmoid activation function.

The gating weights $\mathbf{G}$ are then reshaped into $\mathcal{MG} \in \mathbb{R}^{4\times n}$  to align with the multi-scale recognition results $\mathcal{MC}$. Finally, element-wise multiplication is performed, followed by a summation to produce the final recognition results $\mathcal{MR} \in \mathbb{R}^n$:
\begin{equation}
  \mathcal{MR} = \sum_{i=1}^{4}{\mathcal{MG}_i \cdot \mathcal{MC}_i}.
\end{equation}

The gating weights $\mathcal{MG}$ dynamically control the contribution of information from each scale, leveraging cues from all scales. For a particular scale, the proportion of classification results included in the aggregation depends on the corresponding value in $\mathcal{MG}$. The larger the value, the more ideal the recognition result at that scale. This dynamic selection ensures that the model aggregates only the most appropriate classification results from different scales.

Additionally, bird species exhibit statistical differences in their scale distributions. Our gating mechanism aggregates cues across all scales and finely adjusts the weights of the classification scores on all classes from all scales, effectively addressing this challenge.

\subsection{Loss Functions}
With the output from the gating mechanism, we obtain five recognition results (e.g. $\eta_1,\eta_2,\eta_3,\eta_4,\mathcal{MR}$). For clarity, we use $Y=\{ y_1,y_2,y_3,y_4,y_5\}$ to represent the prediction. For each result $y_i$, we use the cross-entropy loss function with label-smoothing \cite{SzegedyVISW16} for optimization, defined as:
\begin{align}
     \mathcal{L}_s &= \sum_{y \in Y}\sum_{t=1}^{n}-\hat{y}^t_\beta\mathrm{log}y^t ,\\
     \hat{y}_\beta^t&=
\begin{cases}
\beta, &  t =\hat{t}, \\
\frac{1-\beta}{n}, &  t \neq \hat{t} ,
\end{cases}
\end{align}
where $n$ is the number of classes, $t$ denotes the index of the label element, $\hat{t}$ is the index of the ground-truth class, and $\beta \in [0,1]$ is a smoothing factor. We set $\beta$ to incrementally increase in equal intervals of $0.1$, ranging from $0.6$ to $1$. 

To enhance diversity among features and enable multi-scale results to better capture different variations, we incorporate an extra contrastive loss, specifically for $\hat{z_4^0}$ and $\hat{z_3^0}$:
\begin{equation}
  \begin{split}
    \mathcal{L}_{con} = \frac{1}{B^2}\sum_{i=1}^{B}[\sum_{j:y_i=y_j}^{B}(1-\operatorname{cos}(e^i,e^j)) +\\\sum_{j:y_i \neq y_j}^{B}(\operatorname{max}(\operatorname{cos}(e^i,e^j),0)],
  \end{split}
\end{equation}
where $B$ denotes the batch size, $\operatorname{cos}(.,.)$ represents the cosine similarity function, and $e^i$ and $e^j$ refer to $\hat{z_4^0}$ or $\hat{z_3^0}$ extracted from the $i$-th and $j$-th images, respectively. 

The final loss function is defined as:
\begin{align} \label{lossfunction}
    \mathcal{L} = \mathcal{L}_s +\alpha\mathcal{L}_{con},
\end{align}
where $\alpha$ is the weighting factor for $\mathcal{L}_{con}$.

\section{Experiments}
        
\subsection{Datasets}
In our experiment, we demonstrate the effectiveness of our method on three fine-grained recognition datasets: CUB-200-2011 \cite{welinder2010caltech}, NABirds \cite{HornBFHBIPB15}, and the iNat2017 \cite{HornASCSSAPB18}. CUB-200-2011 is a widely used FGVC dataset, consists of $11,788$ images and 200 bird species, dividing $5,994$ images for training and $5,794$ images for testing. NABirds is a larger dataset, consists of $48,562$ images and 555 classes, dividing $23,929$ images for training and $24,633$ images for testing. iNat2017 is a large-scale dataset for fine-grained species recognition, consisting of $859,000$ images from over $5,000$ different species of plants and animals, dividing $579,184$ images for training and $95,986$ images for testing.  

\subsection{Implementation Details}
For all experiments, we use the standard MS-ViT-Base model pre-trained on ImageNet21K as the backbone. Images are first resized to $600\times 600$ pixels and then uniformly cropped to $448\times 448$ pixels. For the training set, we apply random cropping and random horizontal flipping as data augmentation techniques, while for the test set, center cropping is employed. Our training strategy follows prior works~\cite{HeCLKYBW22}, employing an SGD optimizer with a momentum of $0.9$ and a weight decay of $0.0$. 
The batch size is set to $16$ across all datasets. The learning rate is fixed at $0.045$, with the contrastive loss weight set to $0.001$. The hyperparameter $\gamma$ in Eq~\ref{z-score} and $\alpha$ in Eq~\ref{lossfunction} are set as $0.3$ and $0.1$, respectively. We use cosine annealing for learning rate decay during training. All experiments are performed in a CUDA 11.0 environment with PyTorch 1.7.1, utilizing two NVIDIA RTX 3090 GPUs.

\subsection{Main Results}

\begin{table}[t!]
   \centering
    \resizebox{\linewidth}{!}{
    \begin{tabular}{l|c|c}
        \toprule
        Method & Backbone & Acc (\%) \\
        \midrule
        NTS-Net \cite{YangLWHGW18} &\multirow{10}*{ResNet50}    &   87.5 \\ 
        Cross-X \cite{LuoYML0LYL19} & & 87.7 \\
        DCL \cite{WangCLJD20} & & 87.8 \\
        PMG \cite{DuCBXMSG20} & & 89.6 \\
        MCEN \cite{LiWZ21} & & 89.3\\
        GaRD \cite{ZhaoYHL21} & & 89.6\\
        CMN \cite{DengMGZ22} & & 88.2 \\
        MA-ASN \cite{ZhangH022} & & 89.5 \\
        GDSMP-Net \cite{KeCCLG23} &  & 89.9 \\
        TA-CFN \cite{GuanYLZSS23} &  & 90.5 \\
        \midrule
        ViT \cite{DosovitskiyB0WZ21} & \multirow{10}*{ViT-Base} & 90.6 \\
        RAMS-Trans \cite{HuJZHZH021} & &  91.3\\
        AF-Trans \cite{ZhangCZLWLC22} & & 91.6 \\
        DCAL \cite{ZhuKLLTS22} & & 92.0 \\
        TransFG \cite{HeCLKYBW22} &  & 91.7 \\
        SIM-Trans \cite{0002HP22} && 91.8 \\
        IELT \cite{XuWJL23} && 91.8\\
        MpT-Trans \cite{WangFM23} && 92.0 \\
        ACC-ViT \cite{ZhangCWLX24} &  & 91.8 \\
        MP-FGVC \cite{Jiang0GDHL24} & & 91.8 \\
        \midrule
        TransIFC+ \cite{10023961} & \multirow{2}*{Swin-Base} & 91.0 \\
        HERBS \cite{abs-2303-06442} &  & 92.3 \\
         \midrule
        M2Former \cite{abs-2308-02161} & \multirow{2}*{MS-ViT-Base} & 92.4 \\
        \textbf{MDCM (Ours)} &  & \textbf{92.5} \\
        \bottomrule
    \end{tabular}}
    \caption{Comparison with different methods on CUB-200-2011.}
    \label{Result}
\end{table}

\begin{table}[t!]
\centering
\resizebox{\linewidth}{!}{
\begin{tabular}{l|c|c}
\toprule
Method     & Backbone     & Acc(\%)    \\ 
\midrule
Cross-X \cite{LuoYML0LYL19} &\multirow{5}*{ResNet50}& 86.4 \\
PAIRS \cite{GuoF19} && 87.9 \\
GaRD \cite{ZhaoYHL21} && 88.0 \\ 
CMN \cite{DengMGZ22} && 87.8\\
GDSMP-Net \cite{KeCCLG23}    &   & 89.0  \\ 
\midrule
PMG-V2 \cite{DuXMCSG22}    & \multirow{2}*{ResNet101}      & 88.4  \\
MGE-CNN \cite{ZhangHLT19} && 88.6 \\
\midrule
ViT \cite{DosovitskiyB0WZ21} &\multirow{6}*{ViT-Base}& 89.6\\
TransFG  \cite{HeCLKYBW22}   &   & 90.8  \\
IELT \cite{XuWJL23} && 90.8\\                
MpT-Trans \cite{WangFM23}  && 91.3 \\
MP-FGVC \cite{Jiang0GDHL24}       &  &91.0\\
ACC-ViT  \cite{ZhangCWLX24} & &91.4  \\
\midrule
TransIFC+  \cite{10023961} & Swin-Base      & 90.9             \\ 
\midrule
M2Former  \cite{abs-2308-02161} &\multirow{2}*{MS-ViT-Base}  & 91.1  \\
\textbf{MDCM (Ours)}  &   & \textbf{92.0}               \\
\bottomrule
\end{tabular}}
\caption{Comparison with different methods on NABirds.}
\label{ResultonNABirds}
\end{table}

\subsubsection{Results on CUB-200-2011}
The main results on CUB-200-2011 are shown in Table~\ref{Result}. Our method achieves a top-1 accuracy of $92.5\%$, demonstrating a substantial improvement over CNN-based approaches. When compared to ViT-based methods such as TransFG~\cite{HeCLKYBW22}, DCAL~\cite{ZhuKLLTS22} , and M2Former~\cite{abs-2308-02161}, our method outperforms them by $0.8\%$, $0.5\%$, and $0.1\%$, respectively. Furthermore, it achieves a $1.5\%$ improvement over TransIFC~\cite{10023961}, highlighting the effectiveness of our proposed method.

\subsubsection{Results on NABirds}
The main results on the NABirds dataset are shown in Table~\ref{ResultonNABirds}. Our method achieves a top-1 accuracy of $92.0\%$, representing an improvement of at least $0.6\%$. Specifically, compared to recent ViT-based state-of-the-art methods~\cite{ZhangCWLX24,Jiang0GDHL24}, our approach delivers improvements of $0.6\%$ and $1.0\%$, respectively. When compared to the method~\cite{abs-2308-02161} using the same backbone, we observe a $0.9\%$ improvement. Additionally, against another bird recognition-focused method, our approach achieves a $1.1\%$ improvement. These results substantiate the effectiveness of our proposed method.

\begin{table}
\centering
\resizebox{\linewidth}{!}{
  \begin{tabular}{l|c|c}
    \toprule
    Method & Backbone &Acc(\%) \\ 
    \midrule
    TASN \cite{ZhengFZL19} &\multirow{2}*{ResNet50}& 68.2 \\ 
    SRGN \cite{WangWLCOT24} && 73.6\\
    \midrule
    SSN\cite{RecasensKSMT18} &\multirow{2}*{ResNet101}& 65.2 \\
    IARG\cite{HuangL20} &&66.8 \\ 
    \midrule
    RAMS-Trans \cite{HuJZHZH021} &\multirow{6}*{ViT-Base}& 68.5\\
    AF-Trans \cite{ZhangCZLWLC22}  &&68.9 \\
    SIM-Trans \cite{0002HP22} && 69.9\\
    TransFG  \cite{HeCLKYBW22}     &    & 71.7  \\
    MFVT   \cite{lv2022mfvt}       &   & 72.6  \\ 
    ACC-ViT  \cite{ZhangCWLX24}     &     & 77.0 \\ 
    \midrule
    M2Former \cite{abs-2308-02161} &\multirow{2}*{MS-ViT-Base}    & 77.8  \\
    \textbf{MDCM (Ours)} &     & \textbf{79.8}  \\
  \bottomrule
\end{tabular}}
\caption{Comparison with different methods on iNat2017. }
\label{iNat2017} 
\end{table}

\subsubsection{Results on iNat2017}
The main results on the large-scale iNat2017 dataset are shown in Table~\ref{iNat2017}. Our method achieves at least a $2.0\%$ improvement over other approaches, even when some of these methods employ significantly larger backbones~\cite{RyaliHB000ACPHM23,HeCXLDG22}. This highlights the efficiency and effectiveness of our approach.

\begin{table}
\centering
  \label{NabridsResults} 
  \begin{tabular}{lc}
    \toprule
    Mechanisms & Score \\
    \midrule
    Addition Operation   & 113   \\
    Gating Operation   & \textbf{173} \\
  \bottomrule
\end{tabular}
\caption{Impact of different aggregation choices in MSDA.}
\label{Gatingmechanismscore}
\end{table}

\subsection{Ablation Study}
\subsubsection{Effectiveness of Proposed Modules}
We investigate the impact of the proposed modules, with results shown in Table~\ref{Ablationtable}. The baseline MS-ViT achieves a top-1 accuracy of $91.6\%$ on CUB-200-2011. Introducing multi-scale cues from multiple stages (\emph{i.e.,~}Table \ref{Ablationtable}(b)) results in a $0.4\%$ improvement, demonstrating the effectiveness of our token selection design. Notably, for Table~\ref{Ablationtable}(b), the top-1 accuracy corresponds to the best accuracy among the pre-aggregation results, whereas in experiments with MSDA, it refers to the output of MSDA. 
Next, integrating our gating mechanism to dynamically aggregate recognition results from diverse multi-scale cues increases accuracy to $92.3\%$ (\emph{i.e.,~}Table \ref{Ablationtable}(c)). This finding highlights a limitation of prior methods, which process images at different scales across stages and rely on final features for downstream tasks, often failing to meet FGBR’s need for diverse features. By contrast, multi-scale cues learned at different stages provide sufficient diversity. However, correct classifications are dispersed across these multi-scale features. MSDA dynamically aggregates the correct results, yielding a significant performance boost. Finally, applying MSCA to adjust the activation of cues in each stage delivers an additional $0.2\%$ improvement, achieving a top-1 accuracy of $92.5\%$.

\begin{table}
  \centering
    \begin{tabular}{ccccc}
      \toprule
      & MSTS                           & MSDA                                                     & MSCA                           & Acc(\%)  \\
      \midrule
  (a) &                                &                                                                &                                & 91.6 \\
  (b) & \checkmark                     &                                                                &                                & 91.9 \\
  (c) & \multicolumn{1}{c}{\checkmark} & \multicolumn{1}{c}{\checkmark}                                 &                                & 92.3 \\
  (d) & \multicolumn{1}{c}{\checkmark} & \multicolumn{1}{c}{\checkmark} & \multicolumn{1}{c}{\checkmark} & \textbf{92.5} \\
\bottomrule
\end{tabular}
\caption{Ablation study of our MDCM on CUB-200-2011.}
\label{Ablationtable}
\end{table}

\subsubsection{Contributions of MSDA}
We conduct additional experiments to analyze the effectiveness of the gating mechanism. First, we describe the evaluation approach for the gating mechanism. For each picture, five recognition results are generated, resulting in $2^5$ possible combinations of correct or incorrect results. A robust aggregation choice should aim to maintain high accuracy in the aggregated classification result, even when some pre-aggregation results are incorrect. To quantify this, we assign correction scores ranging from $-4$ to $+4$ across the $32$ possible combinations. The correction score for each image is computed as:
\begin{equation}\label{gatingscore}
\text{Score} =
\begin{cases}
\sum_{i=1}^{4}+(y_i\doteq  y), & \text{if } y_5 = y \\
\sum_{i=1}^{4}-(y_i\oplus y), & \text{if } y_5 \neq y 
\end{cases},
\end{equation}
where $y$ is the ground truth label, $\oplus$ denotes the exclusive OR operation, and $\doteq$ denotes the exclusive NOR operation. In brief, when the aggregated result is correct, the score reflects the number of incorrect pre-aggregation results that were corrected, assigning a corresponding positive value. Conversely, a negative score is assigned when the aggregated result is incorrect. This evaluation approach isolates the impact of pre-aggregation results, ensuring that the accuracy improvement is attributable to the aggregation mechanism rather than  better pre-aggregation results.
According to Table~\ref{Gatingmechanismscore}, even a simply summation of recognition results from multi-scale cues learned at different stages yields positive scores, demonstrating that these cues are sufficiently diverse to address the requirements of FGBR. Moreover, our gating mechanism adaptively aggregates these results, effectively mitigating the negative impact of inappropriate cues and achieving higher accuracy.

\begin{figure}[t!]   
  \centering
  \includegraphics[width=\linewidth]{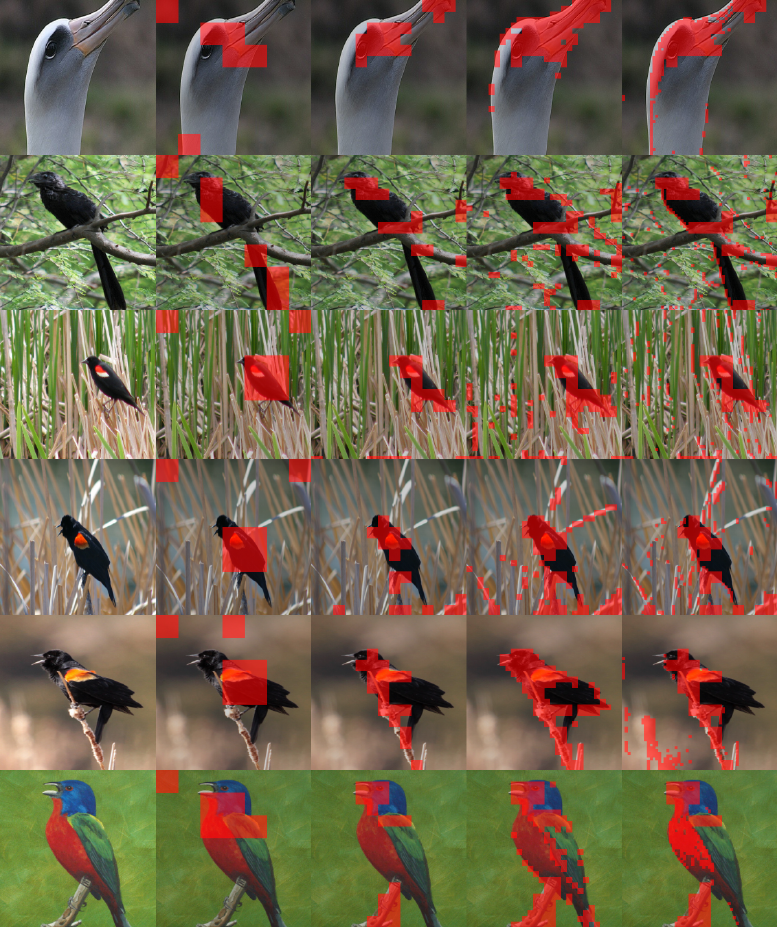}
  \caption{
    The visualization of the MSTS mechanism highlights the selected tokens marked with red rectangles.
    }
      \label{NPSMvisual}
\end{figure}

\subsubsection{Impact of Different Stage Cues}
We further analyze the impact of cues at each stage on recognition accuracy. Following COCO dataset\cite{LinMBHPRDZ14}, we categorize the images in CUB-200-2011 based on bounding box sizes into three groups: \texttt{Small}, \texttt{Medium}, and \texttt{Large}, according to quartiles. Detailed results are presented in Table~\ref{NPSM}. 
The first row shows the performance of the baseline. The second row shows a slight improvement in accuracy, likely due to token selection reducing the influence of noise. Incorporating cues from Stage 3 into the recognition results enhances accuracy for medium and large-scale objects, indicating that Stage 3 cues provide valuable features for objects at these scales. 
Additionally, cues from Stage 2 and Stage 1 improve recognition across all object sizes—small, medium, and large. This suggests that shallow-layer detail information not only aids in identifying small objects but also supplements the recognition of medium and large objects. Overall, capturing diverse multi-scale cues from different stages effectively addresses the challenges of FGBR.

\begin{table}[t!]
  \centering
  \resizebox{\linewidth}{!}{
\begin{tabular}{cccc|ccc|c}
\toprule
\multicolumn{4}{c|}{Cues from different stages} & \multicolumn{4}{c}{Acc(\%)}                                                                 \\ \midrule
Stage $4$  & Stage $3$  & Stage $2$  & Stage $1$ & \multicolumn{1}{c}{\texttt{Large}}  & \multicolumn{1}{c}{\texttt{Medium}}  & \multicolumn{1}{c|}{\texttt{Small}}  & Total \\ \midrule
        &         &         &        & 92.2                     & 91.9                     &90.4 &   91.6    \\
\multicolumn{1}{c}{\checkmark}         &         &         &        & 91.7                    & 92.2                     & 90.5 & 91.7  \\
\multicolumn{1}{c}{\checkmark}         & \multicolumn{1}{c}{\checkmark}         &         &        & 92.5                     & 92.5                    & 90.6 & 92.0  \\
\multicolumn{1}{c}{\checkmark}         & \multicolumn{1}{c}{\checkmark}         & \multicolumn{1}{c}{\checkmark}         &        & \textbf{92.9}                     & 92.7                     & 91.2 & 92.4  \\
\multicolumn{1}{c}{\checkmark}         & \multicolumn{1}{c}{\checkmark}         & \multicolumn{1}{c}{\checkmark}         & \multicolumn{1}{c|}{\checkmark}   & \textbf{92.9} & \textbf{92.8} & \textbf{91.5} & \textbf{92.5}  \\ 
\bottomrule
\end{tabular}}
\caption{The impact of cues learned at different stages}
\vspace{-5mm}
\label{NPSM}
\end{table}

\subsection{Visulization}
Figure~\ref{NPSMvisual} shows the tokens selected at each stage by MSTS, with selected tokens highlighted by red rectangles. The first column displays the original image, while the subsequent columns, from left to right, represent tokens selected from the fourth stage to the first stage. 
In deeper layers, MSTS focuses on primary features at larger scales, whereas in shallower layers, it captures subtle or edge features. For instance, in the third sample, the deeper layers highlight the prominent yellow parts of the wings, while the shallower layers complement this by capturing finer details such as the eyes, beak, and the white feathers beneath the beak. 
Furthermore, with the guidance provided by MSTS, shallower modules effectively model deeper patches at smaller scales, enabling the capture of subtle differences. Overall, MSTS dynamically selects specific tokens across different stages, capturing discriminative representations in both shallow and deep layers, and providing rich cues for final decision-making.

\section{Conclusion}
This paper proposes a novel framework for fine-grained bird recognition, termed Multi-scale Diverse Cues Modeling (MDCM). The proposed framework captures diverse cues at different scales across various stages of a multi-scale Vision Transformer using an ``Activation-Selection-Aggregation'' paradigm. Specifically, the multi-scale cue activation module ensures that the discriminative cues learned at different stages are mutually distinct. Concurrently, a multi-scale token selection module is designed to remove redundant noise and emphasize discriminative, scale-specific cues at each stage. Finally, the selected tokens from each stage are independently utilized for bird recognition, with the recognition results adaptively fused through a multi-scale dynamic aggregation mechanism to make final model decisions. Qualitative and quantitative experiments consistently demonstrate the superiority of our MDCM for the FGBR task.



\bibliography{aaai25}

\end{document}